\pdfoutput=1

\documentclass[11pt]{article}

\usepackage[final]{acl}

\usepackage{times}
\usepackage{latexsym}

\usepackage[T1]{fontenc}

\usepackage[utf8]{inputenc}

\usepackage{microtype}

\usepackage{inconsolata}

\usepackage{graphicx}
\usepackage{amsmath}
\usepackage{caption}
\usepackage{tabularx}
\usepackage{multirow} 
\usepackage{booktabs}
\usepackage[utf8]{inputenc}

\usepackage{amssymb}  
\usepackage{xcolor}
\usepackage{booktabs}
\usepackage{adjustbox}
\title{Emotional Supporters often Use Multiple Strategies in a Single Turn}
\author{
Xin Bai\textsuperscript{1,2,3}, Guanyi Chen\textsuperscript{2,3,4\thanks{Corresponding Authors}}, Tingting He\textsuperscript{2,3,4$^*$}, Chenlian Zhou\textsuperscript{1,2,3},  Yu Liu\textsuperscript{1,2,3}   \\
\textsuperscript{1}Faculty of Artificial Intelligence in Education,\\
\textsuperscript{2}Hubei Provincial Key Laboratory of Artificial Intelligence
and Smart Learning,\\
\textsuperscript{3}National Language Resources Monitor and Research Center
for Network Media,\\
\textsuperscript{4}School of Computer Science, Central China Normal
University \\
\texttt{xin\_b@mails.ccnu.edu.cn, \{g.chen, tthe\}@ccnu.edu.cn}
}

\begin{document}
\maketitle
\begin{abstract}
Emotional Support Conversations (ESC) are crucial for providing empathy, validation, and actionable guidance to individuals in distress. However, existing definitions of the ESC task oversimplify the structure of supportive responses, typically modelling them as single strategy–utterance pairs. Through a detailed corpus analysis of the ESConv dataset, we identify a common yet previously overlooked phenomenon: emotional supporters often employ multiple strategies consecutively within a single turn. We formally redefine the ESC task to account for this, proposing a revised formulation that requires generating the full sequence of strategy–utterance pairs given a dialogue history. To facilitate this refined task, we introduce several modelling approaches, including supervised deep learning models and large language models. Our experiments show that, under this redefined task, state-of-the-art LLMs outperform both supervised models and human supporters. Notably, contrary to some earlier findings, we observe that LLMs frequently ask questions and provide suggestions, demonstrating more holistic support capabilities.
\end{abstract}
\section{Introduction}

Emotional Support Conversations (ESC) play a critical role in helping individuals manage emotional distress and navigate personal challenges through meaningful, empathetic dialogue~\citep{langford1997social,heaney2008social}. The goal of ESC is not only to reduce emotional intensity but also to foster a sense of connection, validation, and clarity during difficult times. Despite its importance, providing effective emotional support remains a complex, nuanced task, often requiring sensitivity, contextual understanding, and advanced interpersonal skills—traits that even humans may struggle to consistently exhibit~\citep{burleson2003emotional}. As such, building intelligent systems capable of engaging in ESC presents both a valuable opportunity and a significant challenge.

\begin{figure*}[htbp] 
    \centering 
    \includegraphics[width=1\textwidth]{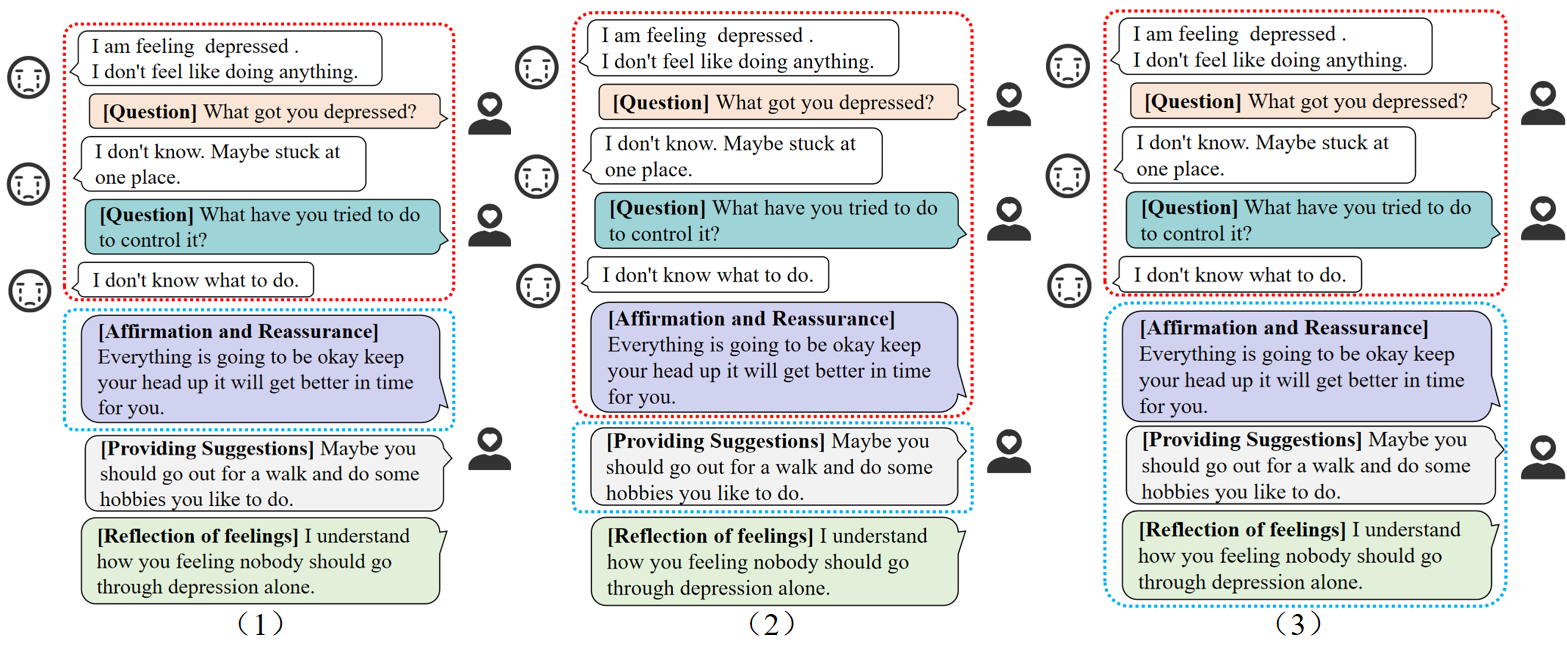} 
    \caption{Example of an emotional support conversation. The part in the red window is the dialogue history, and the part in the blue window is the target that the ESC task asks an ESDS to produce. This Figure contains three tasks on ESC: (1) emotional support utterance generation; (2) emotional support message continuation; (3) the refined ESC task. } 
    \label{fig:example} 
\end{figure*}

To this end, \citet{liu-etal-2021-towards} proposed the ESC framework, which is grounded in Hill’s helping skills theory~\citep{hill1999helping}, and introduced the ESConv dataset. In the ESC framework, an ESC takes place between a help-seeker and an emotional supporter. The framework recommends that supporters follow three stages (i.e., exploration, comforting, and action), and employ a total of eight associated strategies. Figure~\ref{fig:example} presents an example of an ESC. Based on the help-seeker’s inquiry, the supporter begins with the exploration stage, using strategies such as asking questions when essential details about the seeker’s situation are missing. Once the necessary information is obtained, the supporter may provide comfort by reflecting the seeker’s emotions or offering affirmations, and/or take action by suggesting possible next steps.

Based on this framework and the ESConv dataset, \citet{liu-etal-2021-towards} defined the ESC task. Since then, a range of Emotional Support Dialogue Systems (ESDS), built on deep learning techniques or Large Language Models (LLMs), has been developed~\citep{tu-etal-2022-misc,zhao2023transesc,cheng2023pal,zheng-etal-2023-augesc,deng-etal-2023-knowledge,kang-etal-2024-large}. Interestingly, recent studies have found that LLMs, despite their demonstrated effectiveness in various psychological counselling tasks~\citep{ayers2023comparing}, often struggle with the ESC task. Specifically, they tend to select inappropriate strategies~\citep{kang-etal-2024-large} and frequently fail to provide suggestions to seekers~\citep{bai2025holistic}.

We argue that the observed discrepancy in LLM performance stems from a limitation in the current definition of the ESC task. Specifically, prior studies defined the ESC task as predicting the supporter’s next strategy and generating a corresponding utterance based on the dialogue history. However, this formulation overlooks a critical detail: the last utterance in the dialogue history is not necessarily from the supporter. As illustrated in the example ESC in Figure~\ref{fig:example}, a supporter may employ multiple strategies within a single turn, resulting in multiple utterances. We refer to this phenomenon as the Consecutive Use of multiple Strategies (CUS). This observation reveals that the ESC task actually comprises two distinct subtasks. When the last utterance is from the seeker, the task is the \emph{emotional support utterance generation} (Figure~\ref{fig:example}(1)). When the last utterance is from the supporter, the task becomes continuing the supporter’s previous message, i.e., \emph{emotional support message continuation} (Figure~\ref{fig:example}(2)). 

In fact, the second subtask does not occur in isolation in real-world settings, as it presupposes that the decision to continue a prior message has already been made. This presents two key implications. First, when evaluating ESDSes on the ESConv dataset, existing zero-shot or few-shot LLM-based ESDSes fail to understand that they need to distinguish between these subtasks and that they need to generate only a single strategy–utterance pair out of multiple pairs. Second, in real-life deployments, supervised deep learning-based ESDSes typically perform only the first subtask, generating a single strategy–utterance pair in response to each seeker’s message. This behaviour diverges from that of human supporters, who may employ CUS in a turn.

In response to this issue, we argue that modelling CUS is essential for building effective ESDSes and for accurately defining the true ESC task. As illustrated in Figure~\ref{fig:example}(3), the ESC task should be reformulated as predicting all strategies and their corresponding utterances within a single supporter turn, based on a dialogue history in which the last utterance must come from the seeker. In this study, we begin with a brief corpus analysis of the ESConv dataset to examine the prevalence of CUS in ESCs. We then formally redefine the ESC task. Based on this new task formulation, we preprocess the ESConv dataset and introduce several models, ranging from supervised deep learning approaches to LLM-based systems, that explicitly model CUS to accomplish the task. Finally, we evaluate these models using both automatic metrics and human assessments, and analyse their performance with respect to strategy usage and the quality of the emotional support responses they produce. The data and code used in this study are publicly available at: \url{https://github.com/Baixin-ccnu/emotional_support_cus}.

\section{Analysing and Redefining ESC}

We begin by analyzing the ESConv dataset to determine how frequently CUS occurs. Based on our findings, we redefine the ESC task and preprocess the dataset to align with the new task formulation.

\subsection{Analysing ESConv} \label{sec:analyse_esconv}

\begin{figure}
    \centering
    \includegraphics[scale=0.3]{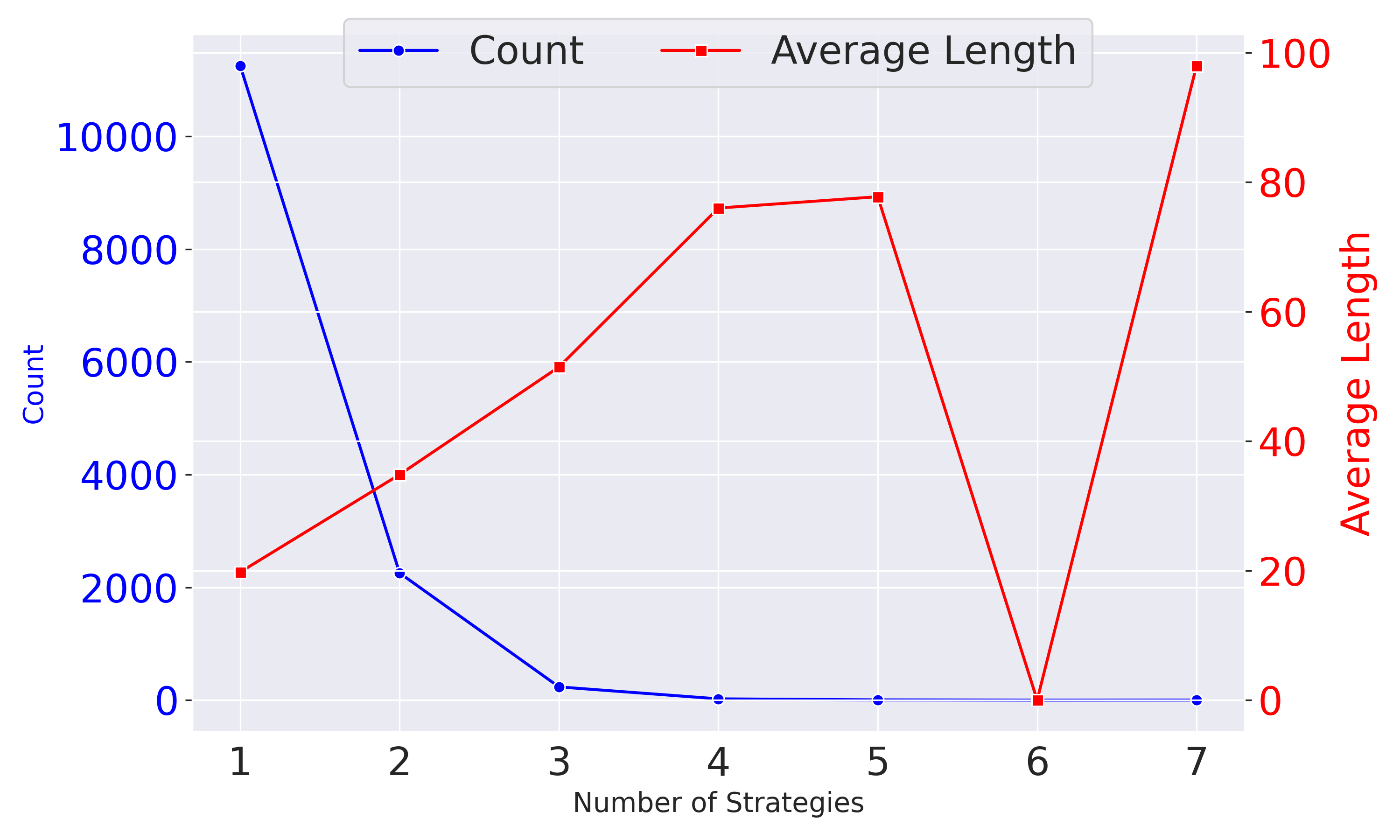}
    \caption{The distribution of responses and the average response length based on the number of strategies employed. Note that the average length is 0 for responses with 6 strategies, as no such responses exist in the ESConv dataset.}
    \label{fig:count}
\end{figure}

To investigate the prevalence of CUS, we count the number of strategies used in each supporter response in the ESConv dataset. Figure~\ref{fig:count} presents the distribution of responses by the number of strategies employed. It also reports the average number of tokens in responses that use different numbers of strategies.

Out of 13,767 responses in ESConv, 2,525 responses use more than one strategy, indicating the presence of CUS. The response with the highest number of strategies employs seven in a single turn. Regarding response length, we observe that when the number of strategies is fewer than four, the response length increases significantly. However, beyond four strategies, the growth in length plateaus. This suggests that when emotional supporters employ a larger number of strategies (i.e., more than four), they begin to shorten each individual utterance within the response.

\subsection{Redefining the ESC Task}

Previously, the ESC task was defined as follows: given a processed dataset $\mathcal{D}=\{c^{(i)}, r^{(i)}\}_{i=1}^N$, where $c^{(i)} = \{ u_1^{(i)}, u_2^{(i)}, \ldots, u_{n_i}^{(i)} \}$ is a sequence of utterances in the dialogue history and $r^{(i)}$ is the target response, the goal is to train an ESDS to predict the target response $r^{(i)}$ based on the dialogue history $c^{(i)}$.

As motivated in the introduction, we refine the task definition by shifting the goal from predicting a single utterance from the supporter to predicting the entire response from the supporter. Formally, given a processed dataset $\mathcal{D} = \{c^{(i)}, r^{(i)}\}_{i=1}^N$, the task is to train an ESDS to predict $r^{(i)}$ based on $c^{(i)}$. Here, $c^{(i)} = \{ u_1^{(i)}, u_2^{(i)}, \ldots, u_{n_i}^{(i)}\}$ denotes a sequence of utterances in the dialogue history, with $u_{n_i}^{(i)}$ being an utterance from the seeker. The target response $r^{(i)} = \{ ur_1^{(i)}, ur_2^{(i)}, \ldots, ur_{m_i}^{(i)} \}$ consists of a sequence of utterances, each associated with a strategy, where $ur^{(i)}$ is an utterance from the supporter. 

\subsection{Pre-precessing ESConv}

Building on the redefined task, we preprocess the ESConv dataset to construct $\mathcal{D}$ for training and evaluating ESDSes. First, we partition the ESConv dataset into training, validation, and test sets using an 8:1:1 ratio. The dataset statistics are presented in Table~\ref{tab:dialogue_stats}. Next, we segment each conversation into data samples in $\mathcal{D}$ by treating each supporter response, which potentially contains multiple strategies and corresponding utterances, as the target $r^{(i)}$, and all preceding utterances as the dialogue history $c^{(i)}$.

In the resulting dataset, each utterance in $r^{(i)}$ is paired with a strategy label. We observed that in some cases, two consecutive utterances within the same response are labelled with the same strategy. While it is possible to enforce a one-to-one mapping between strategies and utterances in $r^{(i)}$, this does not necessarily reflect how human emotional supporters compose their responses. In practice, supporters tend to first select a set of strategies and then generate utterances accordingly~\citep{burleson2003emotional}. As such, repeated use of the same strategy across consecutive utterances may be more appropriately treated as a single strategy instance associated with multiple utterances. 

To account for this, we construct two versions of the dataset. In $D_{v1}$, each utterance-strategy pair is preserved as annotated. In $D_{v2}$, we merge consecutive utterances that share the same strategy and concatenate them, treating them as a single strategy–utterance unit. Maintaining both versions allows us to examine how this difference affects model performance. Since the strategies in each target response appear in a specific order, we refer to them as a strategy sequence. There are 258 distinct strategy sequences in $D_{v1}$ and 193 in $D_{v2}$.

\begin{table}[t]
    \centering
    \resizebox{\columnwidth}{!}{
    \begin{tabular}{lccc}
        \toprule 
        Category & Train & Dev & Test \\
        \midrule 
        \# dialogues & 11024 & 1435 & 1308 \\
        Avg. \# words per utterance & 18.38 & 18.14 & 20.21 \\
        Avg. \# turns per dialogue & 8.52 & 8.59 & 8.38 \\
        Avg. \# words per dialogue & 156.59 & 155.90 & 169.30 \\ 
        \bottomrule 
    \end{tabular}
    }
    \caption{The statistics after partition.}
    \label{tab:dialogue_stats}
\end{table}

\section{Models}

\begin{figure*}[htbp] 
    \centering 
    \includegraphics[width=0.65\textwidth]{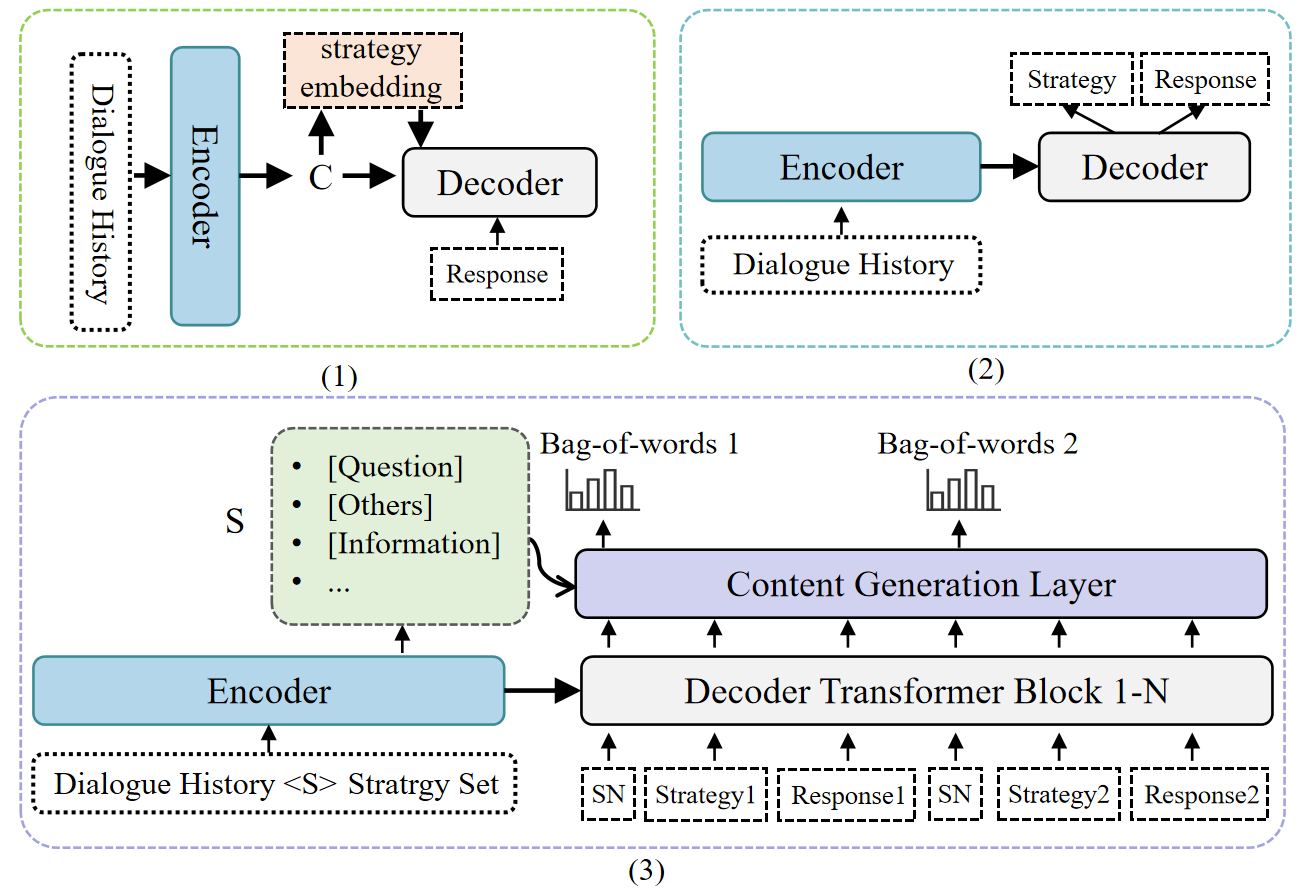} 
    \caption{Diagrams of our supervised approaches. These approaches model CUS as (1) multi-class classification, (2) multi-task learning, and (3) content planning.} 
    \label{fig:model} 
\end{figure*}

We propose four approaches to explicitly model CUS and generate responses accordingly, including three supervised deep learning–based methods and one LLM-based method. Similar to prior works~\citep{liu-etal-2021-towards,zeng-etal-2021-affective,tu-etal-2022-misc,zhao2023transesc}, our supervised approaches are built upon BlenderBot~\citep{roller-etal-2021-recipes}. We formulate the task of modelling CUS through three different paradigms: multi-class classification, multi-task learning, and content planning.

All proposed approaches process the dialogue history in a similar manner. Specifically, they first linearise the dialogue history by inserting special tokens: \texttt{[CLS]} is used to mark the beginning of the history, and \texttt{[EOS]} is inserted after each utterance to indicate its end. The contextualised representation of the dialogue history, denoted as $h$, is then obtained using the encoder of BlenderBot:
\begin{equation}\label{eq:1}
h = \text{Encoder}(\text{[CLS]},u_{1},\text{[EOS]},u_{2},...,u_{n_{i}})
\end{equation}

\subsection{Multi-class Classification}\label{submethod:3.1} 

The first approach treats each distinct strategy sequence in the training set as a separate class, modelling CUS as a multi-class classification task. We refer to this method as \textbf{MCC}. Under this formulation, the model learns to predict the entire sequence of strategies as a single label. When training MCC on $D_{v1}$ or $D_{v2}$, the number of classes corresponds to the number of unique strategy sequences: 258 and 193, respectively.

Concretely, as illustrated in Figure~\ref{fig:model}(1), after obtaining the dialogue history representation $h$ using Equation~\ref{eq:1}, MCC predicts the strategy sequence class $s$ via a fully connected layer followed by a softmax function: 
$$s=\text{Softmax(FCN(h))}.$$
Following~\citet{tu-etal-2022-misc}, MCC employs a strategy embedding layer that maps the predicted class $s$ to its corresponding embedding $E_s$. This embedding, together with the dialogue history representation $h$, is then passed to the decoder of BlenderBot to generate the final response.

The loss function of MCC combines the loss for response generation and the loss for strategy sequence classification. It is defined as:
$$\mathcal{L} = -\sum_{t=1}^{n_r} \log(p(r_t | \mathbf{r}_{j<t}, c)) - \log(p(s | c))$$
where $n_r$ is the length of the response, and $\mathbf{r}_{j<t}$ denotes the previously generated tokens.

\subsection{Multi-task Learning}

The second approach models CUS using a multi-task learning framework, where the model jointly predicts the response and the associated strategies. We refer to this method as \textbf{MTL}. As shown in Figure~\ref{fig:model}(2), the dialogue history representation $h$ is directly passed to the decoder of BlenderBot. During decoding, at each time step, the decoder simultaneously predicts a response token and its corresponding strategy label. This setup allows the model to learn the alignment between content and strategy at a fine-grained level throughout the generation process.

\subsection{Content Planning}

Borrowing an idea from Natural Language Generation (NLG), the third supervised approach models CUS as a content planning task, which is commonly the first step in generating long-form text. We refer to this approach as CP. Unlike the previous two approaches, as shown in Figure~\ref{fig:model}(3), CP slightly modifies the input to help BlenderBot understand what to plan. Specifically, it appends the dialogue history with all candidate strategy words and obtains the input representation in the same way as in Equation~\ref{eq:1}.

During decoding, following~\citet{guan-etal-2021-long} and \citet{li-etal-2021-conversations}, CP introduces a special token \texttt{[SN]}, which not only serves as a representation of the response generated so far but also signals the model to begin planning the next utterance. To implement this, and inspired by~\citet{hu-etal-2022-planet}, CP uses the hidden representation of each \texttt{[SN]} token to predict a bag of words (BOW) that will be used in the upcoming utterance, as an auxiliary task. This task maximises the likelihood of reconstructing the BOWs of the target utterances from the corresponding \texttt{[SN]} representations:
\begin{equation}
\mathcal{L}_{\text{BOW}} = -\frac{1}{J} \sum_{j} \sum_{l} \log p(z_{jl}|\boldsymbol{SN}_j)
\end{equation}
\( J \) denotes the number of target utterances and $p(z_{jl}|\boldsymbol{SN}_j)$ denotes the estimated probability of the $l$-th element in the BOW for the $j$-th target utterance.

Crucially, the strategy for each utterance is predicted as the first token after the corresponding \texttt{[SN]} token, i.e., via standard next-token prediction. This enables the model to treat strategy selection as an integral part of the decoding process, fully embedded in the generation flow. \footnote{We also experimented with predicting strategies explicitly during planning instead of as next tokens (i.e., by using it to replace BOW), but found this reduced performance.} The remaining tokens form the utterance content associated with that strategy. 

The overall training objective combines the generation loss with the auxiliary BOW prediction loss: $\mathcal{L} = \mathcal{L}_{\text{GEN}} + \alpha \mathcal{L}_{\text{BOW}}$, where $\mathcal{L}_{\text{GEN}}$ is the autoregressive generation loss and $\alpha$ is a hyperparameter controlling the weight of the BOW loss.

\subsection{LLM-based ESDS} 

We use the prompt provided in Appendix~\ref{appendix:prompt} to construct ESDS using LLMs. The prompt begins with a task description that includes a list of strategies to choose from. We also tell the LLM that it can select one or more strategies for a single response. Unlike prior work, we omit the definitions of each strategy, as we found their inclusion can degrade performance. The LLM is given two examples before being prompted to generate responses based on the dialogue history.  
\section{Experiments}

\begin{table*}[t]
    \centering
    \small
    \begin{tabular}{lcccccccccc}
        \toprule
\textbf{Model} & \textbf{\#Strategy} & \textbf{EMR} & \textbf{LR} & \textbf{Len.} & \textbf{ALD} & \textbf{D-1} & \textbf{D-2} & \textbf{B-2} & \textbf{B-4} & \textbf{R-L}  \\ \midrule
MCC & 23 & 11.23 & 13.8 & 26.50 & 13.32 & 3.61 & 16.66 & 9.67 & 2.93 & \textbf{18.51} \\
MTL & 754 & 2.21 & 2.9 & 27.07 & \textbf{12.16} & 3.68 & 16.44 & \textbf{10.39} & \textbf{3.52} & 18.09 \\
CP & 61 & \textbf{19.11} & \textbf{24.9} & 18.31 & 13.55 & 4.26 & 27.49 & 5.25 & 1.35 & 15.34 \\
GPT-3.5 & 61 & 13.83 & 17.9 & 42.52 & 18.85 & 4.76 & 25.83 & 6.47 & 1.61 & 16.03 \\
GPT-4o & 73 & 9.25 & 11.7 & 50.40 & 23.21 & 4.97 & 28.61 & 5.80 & 1.40 & 15.16 \\
Qwen-3-8B & 66 & 11.85 & 16.2 & 54.16 & 26.58 & 4.15 & 23.27 & 5.12 & 1.11 & 14.63 \\
Qwen-3-32B & 75 & 5.27 & 6.2 & 53.05 & 25.50 & 4.65 & 28.19 & 5.07 & 0.99 & 14.52 \\
DeepSeek-R1 & 150 & 4.05 & 4.4 & 50.08 & 24.33 & \textbf{7.10} & \textbf{37.42} & 3.88 & 0.71 & 13.36 \\
        \bottomrule
    \end{tabular}
    \caption{Evaluation results in terms of the automatic evaluation metrics on both $D_{v1}$ and $D_{v2}$. `\#Strategy' means the number of distinct strategy sequences used by a model. NB: MTL is unable to be evaluated on $D_{v1}$. }
    \label{tab:result}
\end{table*}

We first describe the experimental settings, and then report model performance on the refined ESC task using both automatic and human evaluations.

\subsection{Backbone LLMs}

To build our LLM-based ESDS, we experimented with a wide range of backbone LLMs, including GPT-3.5~\citep{ouyang2022training}, GPT-4o~\citep{hurst2024gpt}, DeepSeek-R1~\citep{deepseekai2025deepseekr1incentivizingreasoningcapability}, Qwen-3-8B, and Qwen-3-32B~\citep{yang2025qwen3technicalreport}.

\subsection{Evaluation Metrics}

Following prior work on ESDS, we evaluate the quality of generated responses using BLEU-2 (B-2), BLEU-4 (B-4; \citet{papineni2002bleu}), and ROUGE-L (R-L; \citet{lin2004rouge}). To assess diversity, we use Distinct-n (D-n; \citet{li-etal-2016-diversity}), which measures the ratio of unique n-grams in the generated responses.

Additionally, based on the characteristics of the refined ESC task, we propose the following metrics to evaluate both the quality of the model-generated CUS and the corresponding responses.

\paragraph{Exact Match Rate.} Unlike the original ESC task, the revised formulation requires models to predict a sequence of strategies given a dialogue history. The most straightforward way to evaluate the accuracy of the predicted sequence is to check whether it exactly matches the reference sequence. We define the proportion of such exact matches as the Exact Match Rate (EMR).

\paragraph{Levenshtein Ratio.} While EMR provides a strict measure of exact sequence matching, it can be overly restrictive, as it treats all mismatches equally, failing to account for the severity of partial errors. For instance, misclassifying a single strategy in an otherwise correct sequence may be penalised as harshly as predicting an entirely incorrect sequence. To address this limitation, we introduce the Levenshtein Ratio (LR), which quantifies the similarity between the predicted and reference strategy sequences using a normalised edit distance. Specifically, we employ the Levenshtein distance~\citep{levenshtein1966binary} and define LR as:
$$LR = 1 - \frac{\text{Levenshtein Distance}(s_1, s_2)}{\max (len(s_1), len(s_2))},$$
where $s_1$ and $s_2$ denote the predicted and reference strategy sequences, respectively, and $len(\cdot)$ computes the length of a sequence. A higher LR indicates greater similarity between the predicted and reference sequences.

\paragraph{Average Length Difference.}

As shown in Figure~\ref{fig:count}, response lengths vary under the refined ESC task. It is important to examine how the lengths of responses generated by our models compare to those in the reference corpus. To this end, we compute the Average Length Difference (ALD) between generated and reference responses, defined as: 
$$\mathrm{ALD} = \frac{1}{n} \sum_{i = 1}^{n} \left| L_r^{(i)} - L_g^{(i)} \right|,$$ 
where \(n\) is the number of responses, $L_r^{(i)}$ is the length of the $i$-th reference response, and $L_g^{(i)}$ is the length of the $i$-th generated response. This metric provides insight into whether the model tends to over- or under-generate content relative to human responses.

\subsection{Experimental Results}

Table~\ref{tab:result} shows the experimental results on $\mathcal{D}_{v2}$. We report the results on $\mathcal{D}_{v1}$ in Appendix~\ref{sec:appendix_v1} as we found merging consecutive utterances that share the same strategy does not influence the performance of the trained ESDSes. 

Focusing on CUS, MTL produces the largest number of distinct strategy sequences, with a surprising count of 754, followed by DeepSeek-R1 with 150 distinct sequences. However, this diversity comes at a cost: both models exhibit the lowest accuracy in strategy prediction, as reflected by their low EMR and LR scores. In contrast, CP achieves the highest EMR and LR, demonstrating strong capability in modeling CUS through explicit planning. Under the refined ESC task, while LLM-based models still fall short of the best supervised model in strategy prediction, the gap is narrowing compared to previous studies (e.g., \citet{bai2025holistic}). For instance, GPT-3.5 attains the second-highest EMR and LR among all models.

Focusing on the length of the generated responses, we find that, consistent with \citet{bai2025holistic}, LLMs tend to produce longer responses, as they are more likely to employ multiple strategies compared to human supporters (see Section~\ref{sec:analysis} for more details). Consequently, their average length difference (ALD) relative to human responses is also higher.

In terms of response diversity, we observe that LLMs generally outperform supervised models. DeepSeek-R1 achieves the highest Distinct-n scores, with D-1 at 7.1 and D-2 at 37.42. 

However, this increased diversity appears to come at the cost of lower response quality, as reflected in BLEU and ROUGE scores, a trade-off noted in previous work~\citep{zhang-etal-2021-trading}. Interestingly, MTL, despite having the lowest accuracy in strategy sequence prediction, achieves the highest BLEU scores, suggesting that lexical overlap with references may not strongly correlate with accurate strategy planning. It is worth noting that these automatic quality metrics have been shown to have limited validity in evaluating dialogue systems~\citep{liu2016not,reiter2018structured}. Therefore, conclusions about which model produces higher-quality responses should be further validated through human evaluation.

\subsection{Human Evaluation}

\begin{table}[t]
    \centering
    \small
    \resizebox{\columnwidth}{!}{
    \begin{tabular}{llllll}
        \toprule
Model & Flu. & Ident. & Com. & Sug. & O.  \\ \midrule
Human & $5.24^{B,C}$ & $4.76^{C,D,E}$ & $4.38^{B,C}$ & $4.00^{B,C}$ & $4.42^{B}$ \\ \midrule
MCC & $4.92^{C,D}$ & $4.48^{D,E}$ & $4.36^{B,C}$ & $3.68^{C,D}$ & $4.12^{B}$ \\
MTL & $5.70^{A,B}$ & $5.02^{B,C,D}$ & $4.66^{B}$ & $3.74^{C,D}$ & $4.44^{B}$ \\
CP & $4.74^{C,D}$ & $4.10^{E}$ & $3.72^{C}$ & $3.18^{D}$ & $3.80^{B}$ \\
4o & $5.78^{A,B}$ & $5.54^{A,B}$ & $5.44^{A}$ & $4.78^{A,B}$ & $5.28^{A}$ \\
R1 & $\textbf{6.08}^{A}$ & $\textbf{6.06}^{A}$ & $\textbf{5.70}^{A}$ & $\textbf{5.30}^{A}$ & $\textbf{5.60}^{A}$ \\
        \bottomrule
    \end{tabular}}
    \caption{Human evaluation results. `Flu.', `Ident.', `Com.', `Sug.', and `O.' stand for Fluency, Identification, Comforting, Suggestion, and Overall. Per column, results that have \emph{no} superscript letters in common are significantly different from each other ($p < 0.05$).}
    \label{tab:human_result}
\end{table}

We conducted a human evaluation of responses generated by humans and five models trained on $\mathcal{D}_{v2}$: MCC, MTL, CP, GPT-4o, DeepSeek-R1. We randomly sampled 25 dialogue histories from the dataset and collected corresponding responses from each model and the human-generated reference, resulting in a total of $25 \times 6 = 150$ for evaluation. Examples of responses generated by each model are shown in Appendix~\ref{sec:appendix_case}. These items were randomly shuffled before being presented to raters. For the evaluation, we recruited six native-speaking participants with a background in psychology. Following~\citet{liu-etal-2021-towards}, we asked participants to rate each response on a 7-point Likert scale across five dimensions: (1) \textbf{Fluency}: Is the response linguistically smooth, logically coherent, and easy for users to understand? (2) \textbf{Identification}:  Does the response accurately capture the user’s core needs, emotions, or dilemmas?  (3) \textbf{Comforting}: Does it include empathy for the seeker’s emotions (such as acceptance, affirmation, and reassurance)? (4) \textbf{Suggestion}: Does it provide specific and feasible action suggestions for the user’s problem? (5) \textbf{Overall}: In general, do you like this response?

The results of the human evaluation are presented in Table~\ref{tab:human_result}. We conducted pairwise significance testing using the Wilcoxon Signed-Rank Test~\citep{wilcoxon1992individual} with False Discovery Rate correction. Interestingly, and in contrast to the automatic evaluation results, LLMs not only significantly outperform supervised models, but also human supporters across all five evaluation dimensions. Between the two LLMs evaluated, no significant difference was observed. Among the three supervised models, the MTL model, despite its relatively poor performance in predicting accurate strategy sequences (as measured by EMR), achieves the best human evaluation scores. In contrast, the CP model, which excels at predicting strategy sequences, performs the worst. This finding suggests that under the refined ESC task, a given dialogue may correspond to multiple valid strategy sequences. As such, optimising a model solely to replicate the strategy sequences found in the training corpus may be less important than previously assumed.

We also collected qualitative feedback from the participants, who noted that they tended to assign higher scores to responses that appeared not only lengthy but also comprehensive, characteristics commonly observed in responses generated by LLMs. While this highlights a potential advantage of LLMs, it also points to a limitation of the current human evaluation setup, which asks participants to rate isolated responses in isolation. In future work, we plan to design more robust human evaluations, such as those that assess entire emotional support conversations or involve interactive settings.

Our findings diverge from previous studies that constructed ESDSes using LLMs under the original ESC task formulation, which concluded that raw LLMs are inadequate for this task~\citep{kang-etal-2024-large,zhang-etal-2024-escot,bai2025holistic,zhao2025chain}, thereby motivating the use of complex techniques such as Chain-of-Thought (CoT) prompting or fine-tuning. In contrast, we are the first to compare LLM-based ESDSes against both human supporters and supervised models, and we demonstrate that, under the corrected formulation of the ESC task, raw LLMs outperform not only supervised counterparts but even human supporters, at least within corpus-based human evaluation settings. These results suggest that the shortcomings previously attributed to LLMs may actually stem from flaws in the original definition of the ESC task, rather than from inherent limitations of the models themselves. \footnote{NB: Our findings do not invalidate previous LLM-based approaches; rather, under the revised definition of the ESC task, techniques such as CoT prompting or fine-tuning may still provide additional performance gains for LLMs.}


\section{Analysing the Use of Strategy} \label{sec:analysis} 

\begin{figure*}
    \centering
    \includegraphics[scale=0.4]{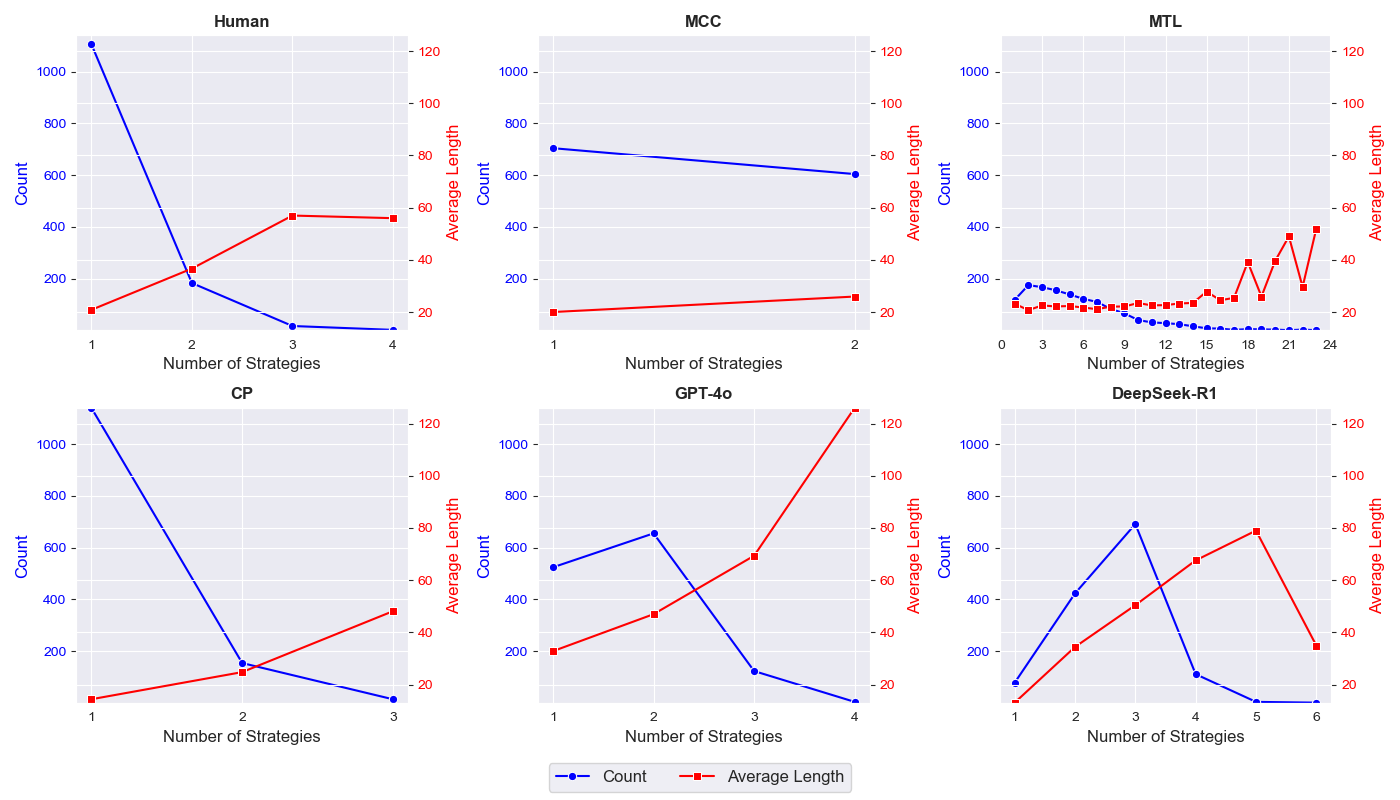}
    \caption{The distribution of responses and the average response length for human, MCC, MTL, CP, GPT-4o, and DeepSeek-R1, based on the number of strategies employed.}
    \label{fig:count_test}
\end{figure*}

\paragraph{Analysing the Consecutive Use of multiple Strategies.} In Section~\ref{sec:analyse_esconv}, we analysed CUS patterns in ESConv. Here, we extend this analysis to the five models evaluated in the human study. The results are presented in Figure~\ref{fig:count_test}, which also includes the test set results from ESConv for reference.

MCC uses at most two strategies per response, which explains its low number of distinct strategy sequences. MTL produces the largest number of distinct strategy sequences and the longest sequences overall, yet its generated responses are not longer than those of the LLMs. This can be attributed to its underlying mechanism for modelling strategy use, where a strategy is predicted at each decoding step. CP's distribution most closely resembles that of human supporters, demonstrating its strong capability in modelling CUS.

As for the two LLM-based systems, compared to human supporters, they are more likely to employ multiple strategies and produce longer responses. DeepSeek-R1 tends to generate longer strategy sequences than GPT-4o, although their average response lengths are similar. Based on the examples in Appendix~\ref{sec:appendix_case}, GPT-4o appears more inclined to associate a single strategy with multiple utterances, whereas DeepSeek-R1 tends to align each strategy with a distinct utterance (potentially reflecting its use of explicit reasoning).

\begin{figure}
    \centering
    \includegraphics[scale=0.32]{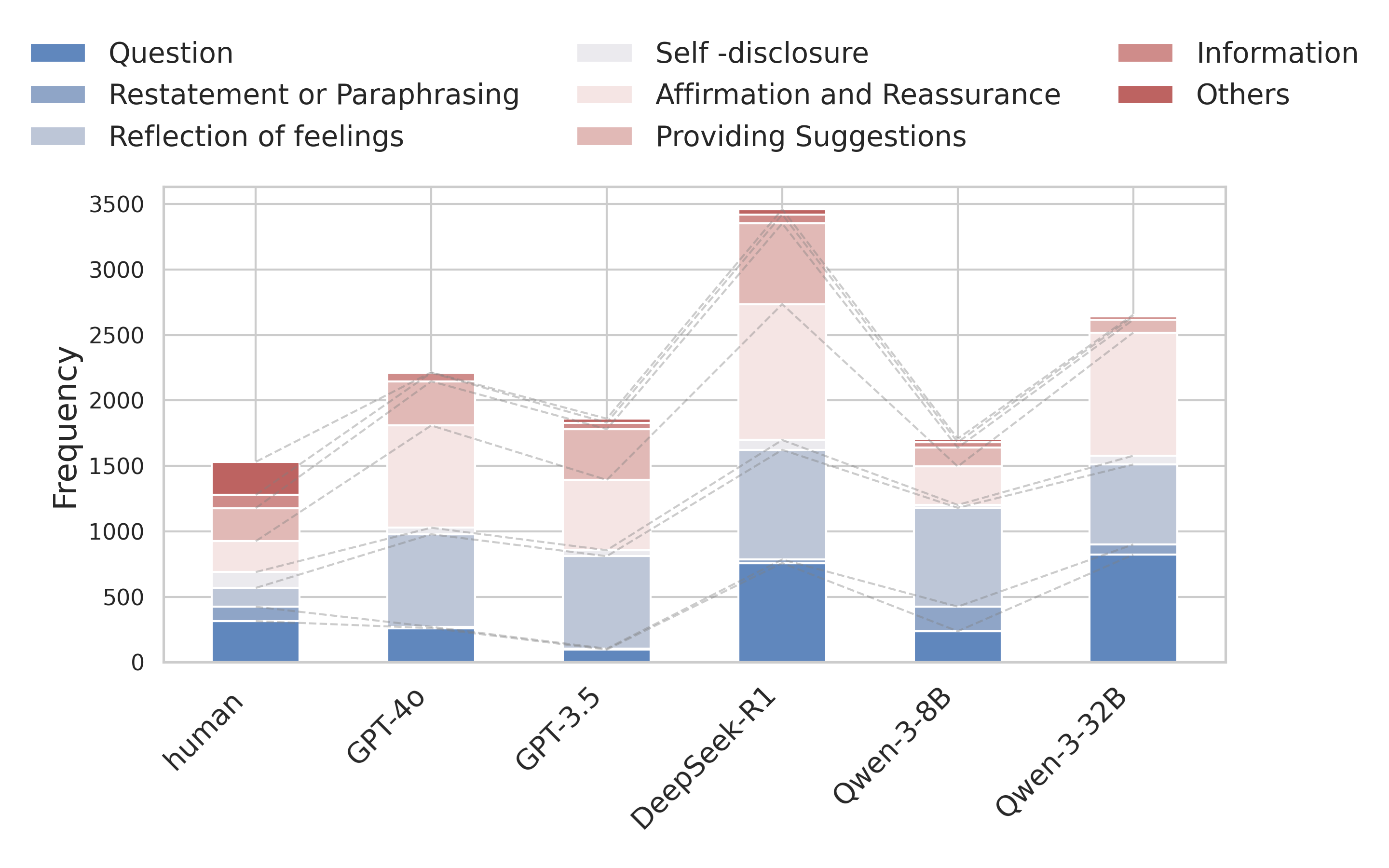}
    \caption{The frequency of each strategy used by each LLM and Human.}
    \label{fig:freq}
\end{figure}

\paragraph{Analysing the Use of Single Strategy.} Previous studies on using LLMs for ESC have found that these models exhibit a preference bias toward certain strategies~\citep{kang-etal-2024-large}, and in particular, they rarely make suggestions~\citep{zhang-etal-2024-escot,bai2025holistic}. Here, we examine whether prompting LLMs to perform the refined ESC task alters this bias. Figure~\ref{fig:freq} presents the frequency of each strategy used by our LLMs. Unlike prior work, we report absolute frequencies rather than normalised distributions, since LLMs are allowed to use multiple strategies within a single response.

The results indicate that while LLMs still exhibit certain strategy preferences, there is no single dominant strategy, unlike what was reported in~\citet{kang-etal-2024-large}. Most LLMs frequently employ [Affirmation and Reassurance], [Reflection of Feeling], [Question], and [Providing Suggestion], thereby covering all three stages of emotional support (exploration, comforting, and action). Notably, under the refined ESC task, LLMs are observed to both ask questions and provide suggestions—strategies they previously avoided~\citep{bai2025holistic}. The best-performing LLM, DeepSeek-R1, even surpasses human supporters in the frequency of both suggesting and questioning. 

The only notable difference is that LLMs almost never use [Other]. This is both explainable and acceptable, as the ESConv dataset assigns the [Other] label to utterances that are difficult to annotate with specific strategies, a classification that may be unclear or inaccessible to LLMs. In our view, the absence of [Other] in LLM-generated responses does not negatively impact response quality.

\section{Conclusion}

In this work, we identify a fundamental limitation in the current formulation of the Emotional Support Conversation (ESC) task. Through a detailed corpus analysis of the ESConv dataset, we uncover the prevalence of the Consecutive Use of multiple Strategies (CUS) within a single supporter turn, an aspect that has been overlooked in prior definitions. We formally refine the ESC task to reflect this more realistic structure of emotional support dialogues, and introduce a revised benchmark aligned with our task definition.

To address this refined task, we propose several approaches, including three supervised deep learning models and a prompt-based LLM method, all explicitly designed to model multi-strategy responses. Evaluation results show that, under this more accurate task setup, state-of-the-art LLMs, GPT-4o and DeepSeek-R1, not only outperform traditional supervised models, but even surpass human supporters in human evaluation. These findings challenge prior conclusions about the limitations of LLMs in ESC, which we argue are rooted in flawed task definitions rather than model inadequacies.

Moreover, we find that, unlike earlier claims, LLMs under our new task formulation frequently engage in asking questions and offering suggestions. This demonstrates that LLMs are more capable emotional supporters than previously understood, provided the task is framed appropriately.

\section*{Limitations}

One limitation of our study lies in the human evaluation setup, which assesses isolated responses rather than full conversations or interactive exchanges. While LLMs outperformed human supporters across all evaluation dimensions, we acknowledge that raters may have favoured longer and more comprehensive responses, an attribute characteristic of LLM outputs. This preference could introduce bias, potentially overestimating the effectiveness of LLMs. Future work should explore more robust evaluation frameworks, including multi-turn dialogue assessments and real-time user interactions, to better capture the nuances of emotional support quality.

Another limitation concerns the profile of our human evaluators. Although all participants had academic backgrounds in psychology, they were not trained psychological counsellors or clinicians. As a result, their judgments may not fully reflect the standards or sensitivities required in professional emotional support contexts. This gap could affect the reliability of the evaluations, especially in dimensions like appropriateness of suggestions or therapeutic impact. Future evaluations would benefit from including certified mental health professionals to provide more expert and clinically informed assessments.

\bibliography{anthology,custom}

\appendix
\setcounter{section}{0}       
\renewcommand{\thesection}{\Alph{section}} 
\section{Strategy Details}\label{appendix:strategy}

\begin{table*}[t] 
\centering
\begin{tabular}{@{}p{\linewidth}@{}}
\toprule
Prompt \\
\midrule
\#\#\# \textbf{TASK DESCRIPTION} \#\#\# \\
Given a two-person dialogue, one person (seeker) expresses his current problem and mood, and the other person (supporter) needs to choose appropriate dialogue strategies (including [Question], [Restatement or Paraphrasing], [Reflection of feelings], [Self - disclosure], [Affirmation and Reassurance], [Providing Suggestions], [Information], [Others]) according to the dialogue content to create an emotional connection with the seeker and provide emotional support, comfort, encouragement, or suggestions.\\
\\
Now, play as the supporter and provide an appropriate response based on the existing context. When responding, you should first specify the strategy or strategies being used, and craft your reply based on those strategies. Note that a single response may involve one or multiple strategies. There is no need to include the thinking process.\\
\\
\#\#\# \textbf{Example} \#\#\# \\
\{example 1\} \\
\{example 2\} \\
\\
\#\#\# \textbf{Dialogue Context} \#\#\# \\
\{context\} \\
\bottomrule
\end{tabular}
\caption{Prompt for response generation with LLMs.}
\label{tab:example_prompts}
\end{table*}

Here, we directly adopted from \citet{liu-etal-2021-towards} to help readers to learn about the specific meaning of each strategy more conveniently.
\paragraph{Question} Asking for information related to the problem to help the help-seeker articulate the issues that they face. Open-ended questions are best, and closed questions can be used to get specific information.
\paragraph{Restatement or Paraphrasing} A simple, more concise rephrasing of the help-seeker’s statements that could help them see their situation more clearly.
\paragraph{Reflection of Feelings} Articulate and describe the help-seeker’s feelings.
\paragraph{Self-disclosure} Divulge similar experiences that you have had or emotions that you share with the help-seeker to express your empathy.
\paragraph{Affirmation and Reassurance} Affirm the helpseeker’s strengths, motivation, and capabilities and provide reassurance and encouragement.
\paragraph{Providing Suggestions} Provide suggestions about how to change, but be careful to not overstep and tell them what to do.
\paragraph{Information} Provide useful information to the help-seeker, for example with data, facts, opinions, resources, or by answering questions.
\paragraph{Others} Exchange pleasantries and use other support strategies that do not fall into the above categories.

\section{Prompt Details}\label{appendix:prompt}
The prompts employed in our experiments are shown in Table 12. To ensure a clear understanding of the task, Task description and strategy are prompted to LLMs. In addition to the dialogue context, we have also included complete dialogue segments, which contain various examples of strategy responses in different quantities.

\section{Results on $\mathbf{D}_{v1}$} \label{sec:appendix_v1}

Table~\ref{tab:result_v2} charts the evaluation results on $\mathcal{D}_{v1}$.

\begin{table*}[t]
    \centering
    \small
    \begin{tabular}{lcccccccccc}
        \toprule
\textbf{Model} & \textbf{\#Strategy} & \textbf{EMR} & \textbf{LR} & \textbf{Len.} & \textbf{ALD} & \textbf{D-1} & \textbf{D-2} & \textbf{B-2} & \textbf{B-4} & \textbf{R-L}  \\ \midrule
MCC & 33 & 9.17 & 11.8 & 27.03 & 13.64 & 3.52 & 16.73 & 9.63 & 2.74 & 18.61 \\ 
CP & 31 & 17.58 & 28.4 & 24.55 & 13.95 & 2.64 & 19.58 & 7.58 & 1.75 & 16.61 \\
GPT-3.5 & 71 & 12.76 & 16.4 & 43.48 & 19.44 & 4.72 & 25.76 & 6.39 & 1.59 & 15.60 \\
GPT-4o & 77 & 7.26 & 9.2 & 57.76 & 28.78 & 4.49 & 26.27 & 5.52 & 1.27 & 15.11 \\
Qwen-3-8B & 68 & 11.69 & 16.1 & 54.34 &  26.68 & 4.15 & 23.28 & 5.19 & 1.10 & 14.63 \\
Qwen-3-32B & 83 & 5.04 & 6.1 & 53.11 & 25.61 & 4.60 & 28.04 & 5.17 & 1.04 & 14.61 \\ 
DeepSeek-R1 & 162 & 3.51 & 4.1 & 50.18 & 24.28 & 7.22 & 37.95 & 4.13 & 0.79 & 13.56 \\
        \bottomrule
    \end{tabular}
    \caption{Evaluation results in terms of the automatic evaluation metrics on $D_{v1}$. NB: MTL is unable to be evaluated on $D_{v2}$. }
    \label{tab:result_v2}
\end{table*}

\section{Case Study} \label{sec:appendix_case}

Table~\ref{tab:model_responses} shows response examples generated by our five models in the human evaluation and humans. 

\begin{table*}[!htbp]
    \centering
    \begin{adjustbox}{max width=\textwidth}
        \small
        \begin{tabular}{@{}p{\linewidth}@{}}
            \toprule
            \textbf{Dialogue History} \\
            \midrule
            \textbf{seeker}: I am getting depressed staying at home. I have not worked in 4 years. \\
            \textbf{supporter}: \emph{[Question]} I understand that can be very isolating. Do you have any family close by or friends that you can talk to? \\
            \textbf{seeker}: I dont have any friends or family here. \\
            supporter: \emph{[Question]} Are you experiencing a problem with finding a job?  \\
            \textbf{seeker}: I have chronic pain issues and i am on disabilty. \\
            \textbf{supporter}: \emph{[Affirmation and Reassurance]} I am sorry and want you to know that reaching out is a great step today. Good for you in the effort. \\
            \textbf{seeker}: How can you help me. \\
            \textbf{supporter}: \emph{[Self - disclosure]} I would like you to know that I was in a car accident and was in a lot of pain for about a year. I was able to work through it but know it is so hard. How were you hurt? \\
            \textbf{seeker}: I got injured at worek. As my hand got trapped in a machine. \\
        \end{tabular}
    \end{adjustbox}

    \vspace{5pt}

    \begin{adjustbox}{max width=\textwidth}
        \small
        \begin{tabular}{@{}p{\linewidth}@{}}
            \toprule
            \textbf{Responses from Models} \\
            \midrule
            \textbf{Human}: \emph{[Reflection of feelings]} Oh, wow! That is really traumatic and I can't imagine that happening to me. You have been through a lot of pain and suffering. I am so sorry. \\
            \textbf{MCC}: \emph{[Affirmation and Reassurance]-[Others]} Oh wow, that is horrible. I am so sorry to hear that. I can understand how you feel. I had a bad experience with a car crash too. I know how you are feeling. I also had a lot more pain in my life than you do. \\
            \textbf{MTL}: \emph{[Reflection of feelings]-[Question]-[Reflection of feelings]-[Question]-[Reflection of feelings]-[Self - disclosure]-[Question]-[Self - disclosure]-[Reflection of feelings]-[Self - disclosure]-[Reflection of feelings]} I am so sorry to hear that. I have had to deal with a lot of pain in my life in the past. I understand how you feel. \\
            \textbf{CP}: \emph{[Restatement or Paraphrasing]} I understand. so you were in an abusive relationship? \\
            \textbf{GPT-4o}: \emph{[Reflection of feelings]} That sounds incredibly painful and traumatic. I can understand how such an injury can impact both your physical and emotional well-being. How are you coping with the chronic pain and the changes in your daily life? \\
            \textbf{Deepseek-R1}: \emph{[Reflection of feelings]-[Affirmation and Reassurance]-[Question]} That sounds incredibly traumatic. I'm so sorry you went through that. Chronic pain on top of such an experience must feel overwhelming. How has your healing journey been since the injury? \\
            \bottomrule
        \end{tabular}
    \end{adjustbox}
    \caption{Example of the responses from our models and Human.} 
    \label{tab:model_responses}
\end{table*}



\end{document}